\documentclass[journal]{IEEEtran}%
\usepackage{amsfonts}
\usepackage{amsmath}
\usepackage{amssymb}
\usepackage{graphicx}
\usepackage{multirow}
\usepackage{amsfonts}
\usepackage{amsmath}
\usepackage{amssymb}
\usepackage{multirow}
\usepackage{graphicx}
\usepackage{mathptmx}
\usepackage{amsfonts}
\usepackage{amsmath}
\usepackage{amssymb}
\usepackage{bm}
\usepackage{multirow}
\usepackage{url}
\usepackage{cite}
\usepackage{color}%
\providecommand{\U}[1]{\protect\rule{.1in}{.1in}}
\ifCLASSINFOpdf
\else
\fi
\begin{document}

\title{A Control Performance Index for Multicopters Under Off-nominal Conditions}
\author{Guang-Xun Du,~Quan Quan,~Zhiyu Xi, Yang Liu and Kai-Yuan Cai \thanks{The
authors are with School of Automation Science and Electrical Engineering,
Beihang University, Beijing 100191, China (dgx@buaa.edu.cn;
qq\_buaa@buaa.edu.cn; z.xi@buaa.edu.cn; qwertyliuyang@buaa.edu.cn;
kycai@buaa.edu.cn)}}
\maketitle

\begin{abstract}
In order to prevent loss of control (LOC) accidents, the real-time control
performance monitoring problem is studied for multicopters. Different from the
existing work, this paper does not try to monitor the performance of the
controllers directly. In turn, the disturbances of multicopters under
off-nominal conditions are estimated to affect a proposed index to tell the
user whether the multicopter will be LOC or not. Firstly, a new degree of
controllability (DoC) will be proposed for multicopters subject to control
constrains and off-nominal conditions. Then a control performance index (CPI)
is defined based on the new DoC to reflect the control performance for
multicopters. Besides, the proposed CPI is applied to a new switching control
framework to guide the control decision of multicopter under off-nominal
conditions. Finally, simulation and experimental results show the
effectiveness of the CPI and the proposed switching control framework.

\end{abstract}

\markboth{Journal of \LaTeX\ Class Files,~Vol.~0, No.~0, August~2017}{Shell
\MakeLowercase{\textit{et al.}}: Bare Demo of IEEEtran.cls for IEEE Journals}






\IEEEpeerreviewmaketitle

\section{Introduction}

\IEEEPARstart{M}{ulticopters }are attracting increasing attention in recent
years \cite{ymz2017multirotor}. The growing interest is partly due to the fact
that multicopters can be used in numerous applications such as surveillance,
inspection, and mapping. Besides, applications of large multicopters are
becoming eye-catching, whilst there exists potential risk in civil safety if
they crash \cite{belcastro2016aircraft}, especially in urban areas. Therefore,
it is of great importance to consider the flight safety problem and prevent
loss of control (LOC) \cite{Belcastro2010,belcastro2010validation} accidents
of multicopters.

Current multicopter autopilots are primarily designed for operation under
nominal conditions (e.g., predefined vehicle weight distribution, good vehicle
health, and acceptable wind disturbances) by the designers. However, it is
unavoidable to use the multicopter under off-nominal conditions
\cite{ymz2017multirotor,Belcastro2010} (e.g., additional payloads,
propulsor\footnote{A propulsor is composed of a propeller, a motor, and an
electronic speed control module, and is powered by a battery. In the
literature, the term \textquotedblleft rotor\textquotedblright,
\textquotedblleft thruster\textquotedblright\ are also used.} degradation, and
unacceptable wind disturbances). Therefore, it is necessary to assess the
performance of the vehicle under off-nominal conditions \cite{IRAC2009}. Based
on the performance assessment results, proper failsafe control actions can be
performed which do not worsen the situation any further. Therefore, a
performance index, which will warn the users or guide the autopilots if the
multicopter is working under off-nominal conditions, is essential.

Control performance monitoring (CPM) is an interesting and important topic.
This can be evidenced by the reviews in
\cite{qin1998control,qin2007recent,bauer2016current} and the references
therein. In \cite{qin1998control}, an overview of the status of control
performance monitoring and assessment using minimum variance principles was
provided. The overview of multiple input multiple output (MIMO) control
performance monitoring was given by \cite{qin2007recent} while the review
\cite{bauer2016current} reported the most recent results in CPM research and
their use in industry. From these literature, one can see that most of the CPM
methods focus on the industrial production processes and try to monitor the
output variance, step changes, settling time, decay ratio, or stability margin
of the control systems online. For aircraft, the authors in
\cite{lichter2009flight} provided a software tool for monitoring control law
stability margins on-line in quasi-real-time. Robust tracking performance was
proposed as a metric for the quantification of the permissible flight envelope
in \cite{pfifer2016quantifying}. In this paper, the CPM problem of multicopter
systems is considered. Here, we do not try to monitor the performance of the
controllers. In turn, the disturbances of multicopters under off-nominal
conditions are estimated to affect a proposed index to tell the user whether
the multicopter will be LOC or not. The salient feature of the proposed method
is the ease-of-use, because only some easy-assess multicopter parameters and
estimated disturbance are needed. This will be introduced in the following sections.

In this paper, a new Control Performance Index (CPI) will be proposed based on
Degree of Controllability (DoC)
\cite{PMuller1972doc,Longman1984doc,Klein1982jgcd,kang2009new}. Here, the DoC
is used to measure the system disturbances while the CPI is used to show
whether the system subject to these disturbances is safe or not. However, the
existing definitions of DoC suffer limited feasibility in practice because of
the following reasons: i) most of these definitions, such as the Grammian
matrix based DoC \cite{PMuller1972doc,kang2009new}, do not consider the
control constraints; ii) the state norm based DoC
\cite{Longman1984doc,Klein1982jgcd} only considers symmetrical control
constraints and is recovery time related. Besides, consider the system
expressed by $\mathbf{\dot{x}}=\mathbf{Ax}+\mathbf{B}\left(  \mathbf{u}%
-\mathbf{d}\right)  $, where $\mathbf{x}\in%
\mathbb{R}
^{n}$, and $\mathbf{u}\in\mathcal{U}\subset%
\mathbb{R}
^{m}$, the control constraint set $\mathcal{U}$ shrinks in the presence of the
external disturbance $\mathbf{d}$. Motivated by these, this paper will define
a new kind of DoC for multicopters based on the Available Control Authority
Index (ACAI) as in \cite{du2015jgcd}. Compared with existing DoCs, the new DoC
has the following advantages: i) it is independent of the recovery time; ii)
it can reflect the effect of the disturbance $\mathbf{d}$; iii) it considers
the control constraints. A new CPI is further defined based on the proposed
DoC to demonstrate the control performance of multicopters subject to
disturbance $\mathbf{d}$. The CPI can tell the user whether the multicopter is
safe or not. Besides, the CPI proposed in this paper can be integrated into
the open source autopilot, such as the Ardupilot project \cite{ardupilot},
without the need for extra sensors.

The major contributions of this paper lie in: i) the definition of the new DoC
for multicopter systems, based on which a new CPI is proposed and used to
monitor the real-time control performance of multicopters working under
off-nominal conditions, ii) the proposed CPI is used in a switching control
framework to guide the control decision of multicopters under off-nominal
conditions. Besides, the maximum ACAI (proposed in our previous work
\cite{du2015jgcd}) of a multicopter is given explicitly and is used to define
the new DoC in this paper. In our previous work \cite{du2015jint}, the
degraded control strategy is switched to by remote pilots on the ground,
whereas this paper provides an index based on which the autopilot can perform
the switching automatically.

In Section II, the dynamic models of multicopter systems are introduced, and
the objective of the paper is provided. In Section III, a preliminary on the
ACAI is given, based on which a new DoC and a new CPI are defined for
multicopters. Besides, a step-by-step procedure is provided to obtain the
proposed CPI. In Section IV, the proposed CPI is applied to a switching
control framework. Finally, the effectiveness of the new CPI is demonstrated
by numerical and experimental results in Section V. Section VI concludes the paper.

\section{Problem Formulation}

\subsection{Mathematical Model of Multicopters}

Consider a multicopter consists of a rigid frame equipped with $n_{P}$
propellers. In practice, the multicopter uses the $n_{P}$ propellers to
produce the total thrust denoted by $u_{t}$ and control torques denoted by
$\mathbf{u}_{\tau}\triangleq\lbrack\tau_{x}$ $\tau_{y}$ $\tau_{z}]{^{T}}%
\in{\mathbb{R}^{3}}$ (where $\tau_{x},\tau_{y},\tau_{z}$ are the airframe
roll, pitch and yaw torque of multicopter, respectively). Let $\mathcal{I}%
=\left\{  \mathbf{e}_{x},\mathbf{e}_{y},\mathbf{e}_{z}\right\}  $ denote an
right-hand inertial frame and $\mathcal{A}=\left\{  \mathbf{e}_{1}%
,\mathbf{e}_{2},\mathbf{e}_{3}\right\}  $ denote a (right-hand) body fixed
frame rigidly attached to the aircraft where the center of gravity (CoG) of
the multicopter is chosen as the origin of $\mathcal{A}$. According to
\cite{du2015jgcd} and \cite{schneider2011fault}, the mapping from the
propulsor thrust ${f_{i}},i=1,\cdots,{n_{P}}$ to the thrust and torques vector
$\mathbf{u}_{f}\triangleq\lbrack u_{t}$ $\mathbf{u}_{\tau}^{T}]^{T}$ is given
by $\mathbf{u}_{f}={{\mathbf{B}}_{f}}{\mathbf{f}}$ where ${\mathbf{f}%
}\triangleq\left[  {f_{1}}\text{ }{f_{2}}\text{ }\cdots\text{ }{f_{{n_{P}}}%
}\right]  ^{T}$, ${{\mathbf{B}}_{f}}\in{\mathbb{R}^{4\times{n_{P}}}}$ is the
control effectiveness matrix. Then, by ignoring the aerodynamic damping and
stiffness, the rigid body equations of motion of the multicopter are given by
\cite{tayebi2006attitude}\cite{quan2017introduction}%
\begin{equation}%
\begin{array}
[c]{rll}%
\mathbf{\dot{p}} & = & \mathbf{v}\\
m_{a}\mathbf{\dot{v}} & = & m_{a}g\mathbf{e}_{3}-u_{t}\mathbf{R}_{e_{3}%
}+\mathbf{d}_{v}\\
\boldsymbol{\dot{\Theta}} & = & \mathbf{W}\mathbb{\cdot}\boldsymbol{\omega}\\
\mathbf{J}\boldsymbol{\dot{\omega}} & = & -\boldsymbol{\omega}\times
\mathbf{J}\boldsymbol{\omega}+\mathbf{G}_{a}+\mathbf{u}_{\tau}+\mathbf{d}%
_{\omega}%
\end{array}
\label{nonlinear}%
\end{equation}
where $m_{a}$ denotes the mass of the vehicle, $g$ denotes the acceleration
due to gravity, $\mathbf{e}_{3}\triangleq\lbrack0$ $0$ $1]^{T}$ denotes the
unit vector in $\mathcal{A}$, $\mathbf{G}_{a}$ represents the gyroscopic
torques. The vector $\mathbf{p}\triangleq\lbrack x$ $y$ $h]^{T}$ and
$\mathbf{v}\triangleq\lbrack v_{x}$ $v_{y}$ $v_{h}]^{T}$ denotes the position
and linear velocity of the origin of $\mathcal{A}$ with respect to
$\mathcal{I}$, respectively. The vector $\boldsymbol{\omega}\triangleq\lbrack
p$ $q$ $r]{^{T}}\in{\mathbb{R}^{3}}$, where $p,q,r$ denote the roll, pitch and
yaw angular velocities of the multicopter expressed in $\mathcal{A}$. The
vector $\boldsymbol{\Theta}\triangleq\lbrack\phi$ $\theta$ $\psi]{^{T}}%
\in{\mathbb{R}^{3}}$ where $\phi,\theta,\psi$ denote the roll, pitch and yaw
angles of $\mathcal{A}$ with respect to $\mathcal{I}$. The matrix
${{\mathbf{J}}\triangleq\text{diag}}\left(  {{J_{x}},{J_{y}},{J_{z}}}\right)
\in{\mathbb{R}^{3\times3}}$ is the constant inertia matrix where $J_{x}%
,J_{y},J_{z}$ are the moment of inertia around the roll, pitch and yaw axes of
the multicopter frame, respectively. The matrix $\mathbf{R}_{e_{3}}$ and
$\mathbf{W}$ are given as follows%
\[
\mathbf{R}_{e_{3}}=\left[
\begin{array}
[c]{c}%
s\psi s\phi+c\psi s\theta c\phi\\
-c\psi s\phi+c\phi s\theta s\psi\\
c\theta c\phi
\end{array}
\right]  ,\mathbf{W}=\left[
\begin{array}
[c]{ccc}%
1 & s\phi s\theta/c\theta & c\phi s\theta/c\theta\\
0 & c\phi & -s\phi\\
0 & s\phi/c\theta & c\phi/c\theta
\end{array}
\right]
\]
where $c$ and $s$ are shorthand forms for cosine and sine. The terms
$\mathbf{d}_{v}\triangleq\left[  d_{x}\text{ }d_{y}\text{ }d_{v_{h}}\right]
^{T}$ and $\mathbf{d}_{\omega}\in{{\mathbb{R}}^{3}}$ are used to denote the
unknown disturbances, which cover additional payloads, propulsor degradation,
external disturbances, and unmodeled dynamics. Additional payloads will change
the mass $m_{a}$ and inertia ${{\mathbf{J}}}$ of the multicopter and the
effect can be lumped into $d_{v_{h}}$ and $\mathbf{d}_{\omega}$. The propulsor
degradation will change the control effectiveness matrix ${{\mathbf{B}}_{f}}$
to $\mathbf{E}\triangleq{{\mathbf{B}}_{f}}\left(  {{\mathbf{I}}_{{{n}_{P}}}%
}-\boldsymbol{\Gamma}\right)  \in{{\mathbb{R}}^{4\times{{n}_{P}}}}$, where
$\boldsymbol{\Gamma}\triangleq{\text{diag}}\left(  {{\eta_{1}},\cdots
,{\eta_{{n_{P}}}}}\right)  \in{\mathbb{R}^{{n_{P}}\times{n_{P}}}}$ and
${{\eta}_{i}}\in\left[  0,1\right]  ,i=1,\cdots,{{n}_{P}}$ is used to account
for propulsor efficiency degradation, i.e., that the effectiveness of
propulsor $i$ is reduced by $100{{\eta}_{i}\%}$. If the $i$th propulsor
completely fails, then ${{\eta}_{i}}=1$. Then, propulsor degradation will
introduce the term ${{\mathbf{B}}_{f}}\boldsymbol{\Gamma}\mathbf{f}$ which can
be lumped into $d_{v_{h}}$ and $\mathbf{d}_{\omega}$. As the wind disturbance
only affect the dynamics of multicopters but not the multicopter kinematics,
the wind disturbances can be lumped into $\mathbf{d}_{v}$ and $\mathbf{d}%
_{\omega}$.

From (\ref{nonlinear}), the multicopter system is a typical nonlinear system.
This makes the analysis and design complex. To simplify the model, the
following assumption is used:

\textbf{Assumption 1}. $\sin\phi\approx\phi,\cos\phi\approx1,\sin\theta
\approx\theta,\cos\theta\approx1$ and $u_{t}\approx m_{a}g.$

According to \emph{Assumption 1}, the pitch and roll angles are small, and the
total thrust approximates to the weight of the multicopter. Then, the matrix
$\mathbf{W}$ is approximated to the identity matrix $\mathbf{I}_{3}$ and the
matrix $\mathbf{R}_{e_{3}}$ becomes%
\[
\mathbf{R}_{e_{3}}\approx\left[
\begin{array}
[c]{c}%
\phi\sin\psi+\theta\cos\psi\\
-\phi\cos\psi+\theta\sin\psi\\
1
\end{array}
\right]  .
\]
Based on (\ref{nonlinear}), by lumping the nonlinear term $-\boldsymbol{\omega
}\times\mathbf{J}\boldsymbol{\omega}+\mathbf{G}_{a}$ with $\mathbf{d}_{\omega
}$, the simplified multicopter model is given by%
\begin{align}
&
\begin{array}
[c]{ccl}%
\mathbf{\dot{p}}_{l} & = & \mathbf{v}_{l}\\
m_{a}\mathbf{\dot{v}}_{l} & = & -m_{a}g\mathbf{A}_{\psi}\boldsymbol{\Theta
}_{l}+\mathbf{d}_{l}%
\end{array}
\label{horizental}\\
&
\begin{array}
[c]{ccl}%
\dot{h} & = & v_{h}\\
m_{a}\dot{v}_{h} & = & -u_{t}+d_{h}%
\end{array}
\label{vertical}\\
&
\begin{array}
[c]{ccl}%
\boldsymbol{\dot{\Theta}} & = & \boldsymbol{\omega}\\
\mathbf{J}\boldsymbol{\dot{\omega}} & = & \mathbf{u}_{\tau}+\mathbf{d}_{\tau}%
\end{array}
\label{attitude}%
\end{align}
where $\mathbf{p}_{l}\triangleq\lbrack x$ $y]^{T}$, $\mathbf{v}_{l}%
\triangleq\lbrack v_{x}$ $v_{y}]^{T}$, $\boldsymbol{\Theta}_{l}\triangleq
\lbrack\phi$ $\theta]^{T}$, $\mathbf{d}_{l}\triangleq\left[  d_{x}\text{
}d_{y}\right]  ^{T}$, $d_{h}=d_{v_{h}}+m_{a}g$, $\mathbf{d}_{\tau}%
\triangleq\lbrack d_{l}$ $d_{m}$ $d_{n}]^{T}=\mathbf{d}_{\omega}%
-\boldsymbol{\omega}\times\mathbf{J}\boldsymbol{\omega}+\mathbf{G}_{a}$, and
$\mathbf{A}_{\psi}\triangleq\left[
\begin{array}
[c]{cc}%
\sin\psi & \cos\psi\\
-\cos\psi & \sin\psi
\end{array}
\right]  $.

From (\ref{horizental}), the lateral position $\mathbf{p}_{l}$ is practically
controlled by adjusting roll and pitch angles to desired value $\phi_{c}$ and
$\theta_{c}$ while the yaw angle $\psi$ is controlled to a fixed value
$\psi_{c}$. From (\ref{vertical}) and (\ref{attitude}), the altitude $h$ and
attitude $\boldsymbol{\Theta}$ are controlled by using the $n_{P}$ propellers
to produce the total thrust $u_{t}$ and control torques $\mathbf{u}_{\tau}$.
To distinguish from the lateral dynamics shown by (\ref{horizental}), the
altitude and attitude dynamics shown by (\ref{vertical}) and (\ref{attitude})
are called the \emph{basic dynamics} of the multicopter. Furthermore, the
following assumption is used:

\textbf{Assumption 2}. The desired roll, pitch, and yaw angles $\phi_{c}$,
$\theta_{c}$, $\psi_{c}$ are constants, and the desired roll, pitch, and yaw
angular velocities $p_{c}$, $q_{c}$, $r_{c}$ are zeros.

Then, the multicopter dynamics can be formulated into the following form%
\begin{align}
\mathbf{\dot{x}}  &  =\mathbf{Ax}+\mathbf{B}\left(  \mathbf{u}-\mathbf{d}%
\right)  ,\mathbf{u=H}\boldsymbol{\mu}\label{distrubance_model}\\
\mathbf{A}  &  =\left[
\begin{array}
[c]{cc}%
\mathbf{0}_{n\times n} & \mathbf{I}_{n}\\
\mathbf{0}_{n\times n} & \mathbf{0}_{n\times n}%
\end{array}
\right]  ,\mathbf{B}=\left[
\begin{array}
[c]{c}%
\mathbf{0}_{n\times n}\\
\mathbf{M}%
\end{array}
\right]  \label{ABM}%
\end{align}
where $\mathbf{x\in%
\mathbb{R}
}^{2n}$, $\boldsymbol{\mu}\in\mathcal{U}\subset\mathbf{%
\mathbb{R}
}^{m}$, $\mathbf{u}\in\Omega\subset\mathbf{%
\mathbb{R}
}^{n}$, $\mathbf{M\in%
\mathbb{R}
}^{n\times n}$, $\mathbf{H\in%
\mathbb{R}
}^{n\times m}$. The system (\ref{distrubance_model}) is a linear one with
constrained control inputs. For lateral dynamics, one has%
\begin{align*}
n  &  =2,m=2,\mathbf{x}\triangleq\left[  x\text{ }y\text{ }v_{x}\text{ }%
v_{y}\right]  ^{T},\\
\boldsymbol{\mu}  &  \triangleq\left[  \phi_{c}\text{ }\theta_{c}\right]
^{T},\mathbf{d\triangleq d}_{l},\mathbf{H}\triangleq m_{a}g\mathbf{A}_{\psi
},\mathbf{M\triangleq-}\frac{1}{m_{a}}\mathbf{I}_{n},\\
\mathcal{U}  &  \triangleq\left\{  \boldsymbol{\mu}{|}\phi_{c}\in{\left[
{-\phi}_{\max}{,}\phi_{\max}\right]  ,}\theta_{c}\in{\left[  {-}\theta_{\max
}{,}\theta_{\max}\right]  }\right\}  .
\end{align*}
On the other hand, for basic dynamics, one has%
\begin{align*}
n  &  =4,m=n_{P},\boldsymbol{\mu}\triangleq{\mathbf{f,}}\mathbf{d\triangleq
\lbrack}d_{h}\text{ }-\mathbf{d}_{\tau}^{T}\mathbf{]}^{T},\\
\mathbf{x}  &  \triangleq\left[  h\text{ }\phi-\phi_{c}\text{ }\theta
-\theta_{c}\text{ }\psi-\psi_{c}\text{ }v_{h}\text{ }p\text{ }q\text{
}r\right]  ^{T},\\
\mathbf{H}  &  \mathbf{\triangleq B}_{f},\mathbf{M}\triangleq\mathrm{diag}%
\left(  -m_{a},{{J_{x}},{J_{y}},{J_{z}}}\right)  .
\end{align*}
In practice, ${f_{i}}\in\left[  {0,{K_{i}}}\right]  ,i=1,\cdots,{n_{P}}$
(where $K_{i}$ is the maximum thrust of the $i$-th propulsor) because the
propulsors can only provide unidirectional thrust (upward or downward). As a
result, $\mathcal{U}$ of the basic dynamics is given by $\boldsymbol{\mu}%
\in\mathcal{U}\triangleq\left\{  {\mathbf{f}|}f_{i}\in{\left[  {0,{K_{i}}%
}\right]  ,i=}1,\cdots,{n_{P}}\right\}  $. For both the lateral dynamics and
the basic dynamics, one has $\mathbf{u}\in\Omega\triangleq\left\{
\mathbf{u}|\mathbf{u}=\mathbf{H}\boldsymbol{\mu}\right\}  \subset{{\mathbb{R}%
}^{n}}.$

\subsection{Objective of the Paper}

In practice, the system (\ref{distrubance_model}) is usually controlled by the
controllers given by $\mathbf{u}=\mathbf{u}\left(  \mathbf{x},t\right)  $
designed under nominal conditions. The objective of this paper is to solve the
following problems: i) how to monitor the control performance of the closed
loop system formed by combining system (\ref{distrubance_model}) and
controller $\mathbf{u}=\mathbf{u}\left(  \mathbf{x},t\right)  $, and ii) how
to guide the users or the autopilots based on the monitoring results to keep
multicopters safe under severe off-nominal conditions.

\section{A Control Performance Index For Multicopters}

This section will propose a control performance index for the system in
(\ref{distrubance_model}) and the results will apply to both the lateral
dynamics and basic dynamics of the considered multicopters.

\subsection{Preliminaries}

In practice, if the unknown disturbance $\mathbf{d}$ caused by the off-nominal
conditions makes the system in (\ref{distrubance_model}) uncontrollable, then
the multicopter will be LOC. In order to test the controllability of the
disturbance driven system in (\ref{distrubance_model}), the ACAI based
controllability analysis method in \cite{du2015jgcd} is used. The ACAI,
spurred by the research in \cite{schneider2011fault}, was first proposed in
\cite{du2015jgcd} and originally used to check on the positive controllability
of multicopters. In this paper, the ACAI is extended for system
(\ref{distrubance_model}) and is defined as%
\begin{equation}
\rho\left(  \boldsymbol{\alpha},\partial\Omega\right)  \triangleq\left\{
\begin{array}
[c]{c}%
\min\left\{  \left\Vert \boldsymbol{\alpha-\beta}\right\Vert
:\boldsymbol{\alpha}\in\Omega,\boldsymbol{\beta}\in\partial\Omega\right\} \\
-\min\left\{  \left\Vert \boldsymbol{\alpha-\beta}\right\Vert
:\boldsymbol{\alpha}\in\Omega^{C},\boldsymbol{\beta}\in\partial\Omega\right\}
\end{array}
\right.  . \label{22}%
\end{equation}
Here, $\rho\left(  \boldsymbol{\alpha},\partial\Omega\right)  $ represents the
distance from $\boldsymbol{\alpha}$ to $\partial\Omega$, where $\partial
\Omega$ is the boundary of $\Omega$, ${{\Omega}^{C}}$ is the complementary set
of $\Omega$. Then the ACAI of the system (\ref{distrubance_model}) is defined
as $\rho\left(  \mathbf{d},\partial\Omega\right)  \in\mathbb{R}$ and the
following theorem is obtained directly according to the results in
\cite{du2015jgcd}.

\textbf{Theorem 1.} The system in (\ref{distrubance_model}) is controllable if
and only if $\rho\left(  \mathbf{d},\partial\Omega\right)  >0.$

Physically, $\rho\left(  \mathbf{d},\partial\Omega\right)  $ is the radius of
the biggest enclosed sphere centered at $\mathbf{d}$ in the attainable control
set $\Omega$. The larger the value of $\rho\left(  \mathbf{d},\partial
\Omega\right)  $ is, the larger is the attainable control set. Then the system
has more control margin to reject disturbances. In particular, if $\rho\left(
\mathbf{d},\partial\Omega\right)  $ is zero, no enough control can be provided
to stabilize the system, and the system is therefore LOC. In order to compute
the value of $\rho\left(  \mathbf{d},\partial\Omega\right)  $, a step-by-step
ACAI computing procedure is given in \cite{du2015jgcd}, and the readers are
referred to the toolbox provided in \cite{du2016toolbox} which can be used for
system (\ref{distrubance_model}) after minor modifications.

\subsection{A Control Performance Index}

As mentioned above, the ACAI $\rho\left(  \mathbf{d},\partial\Omega\right)  $
can be used to indicate the largest toleration to disturbances for a
multicopter. However, it is not a control performance index intuitively,
because it does not take the controller into account. To account for this, a
new DoC is defined first, based on which a CPI is proposed.

\subsubsection{A New Definition of the Degree of Controllability}

In this subsection, a virtual ACAI $\rho\left(  \mathbf{u}{_{c}}%
,\partial\Omega\right)  $ is used to normalize the ACAI, where $\mathbf{u}%
{_{c}}=\mathbf{H}\boldsymbol{\mu}_{c}$ is the center of $\Omega$ and
$\boldsymbol{\mu}_{c}={\left[
\begin{array}
[c]{cc}%
0 & 0
\end{array}
\right]  ^{T}}$ for the lateral dynamics while $\boldsymbol{\mu}_{c}=\frac
{1}{2}{\left[
\begin{array}
[c]{cccc}%
{{K_{1}}} & {{K_{2}}} & \cdots & {{K_{{n_{P}}}}}%
\end{array}
\right]  ^{T}}$ for the basic dynamics. According to (\ref{22}), the following
lemma is obtained.

\textbf{Lemma 1}. $\rho\left(  \mathbf{u}{_{c}},\partial\Omega\right)  $ is
the maximum ACAI of system (\ref{distrubance_model}).

\emph{Proof}. In the following, it is assumed that $\rho\left(  \mathbf{d}%
,\partial\Omega\right)  >0$. According to \emph{Theorem 3} in
\cite{du2015jgcd}, if rank$\left(  \mathbf{H}\right)  =n$, then the ACAI
$\rho\left(  \mathbf{d},\partial\Omega\right)  $ is given by
\begin{equation}
\rho\left(  \mathbf{d},\partial\Omega\right)  =\min\left(  {{d}_{1}},{{d}_{2}%
},\cdots,{{d}_{{{s}_{m}}}}\right)  \label{62}%
\end{equation}
where ${{d}_{j}}=+\infty$ if rank$\left(  \mathbf{H}{_{1,j}}\right)  <n-1$
and
\begin{equation}
{{d}_{j}}=\frac{1}{2}\text{sign}\left(  \mathbf{\xi}_{j}^{T}{{\mathbf{H}%
}_{2,j}}\right)  \boldsymbol{\Lambda}{_{j}}{{\left(  \mathbf{\xi}_{j}%
^{T}{{\mathbf{H}}_{2,j}}\right)  }^{T}}-\left\vert \mathbf{\xi}_{j}^{T}\left(
\mathbf{u}{_{c}}-\mathbf{d}\right)  \right\vert \label{63}%
\end{equation}
if rank$\left(  {{\mathbf{H}}_{1,j}}\right)  =n-1$. Here, the matrices
{{$\mathbf{H}$}}${_{1,j}}\in{{\mathbb{R}}^{n\times\left(  n-1\right)  }}$ and
{{$\mathbf{H}$}}${_{2,j}}\in{{\mathbb{R}}^{n\times\left(  m+1-n\right)  }}$
are composed of arbitrary $n-1$ columns and the remaining $m+1-n$ columns of
$\mathbf{H}$, respectively. There are totally ${{s}_{m}}$ cases of
{{$\mathbf{H}$}}${_{1,j}}$ and {{$\mathbf{H}$}}${_{2,j}}$ where ${{s}_{m}%
}=\frac{m!}{\left(  m+1-n\right)  !\left(  n-1\right)  !}.$The vector
${{\mathbf{\xi}}_{j}}\in{{\mathbb{R}}^{n-1}}$ satisfies $\mathbf{\xi}_{j}^{T}%
${{$\mathbf{H}$}}${_{1,j}}=0$,$\ \left\Vert {{\mathbf{\xi}}_{j}}\right\Vert
=1$.

Similarly, if rank$\left(  {{\mathbf{H}}}\right)  =n$, then the virtual ACAI
$\rho\left(  \mathbf{u}{_{c}},\partial\Omega\right)  $ is given by
\begin{equation}
\rho\left(  \mathbf{u}{_{c}},\partial\Omega\right)  =\min\left(  d_{1}%
^{c},d_{2}^{c},\cdots,d_{{{s}_{m}}}^{c}\right)  \label{64}%
\end{equation}
where $d_{j}^{c}=+\infty$ if rank$\left(  {{\mathbf{H}}_{1,j}}\right)  <n-1$
and
\begin{equation}
d_{j}^{c}=\frac{1}{2}\text{sign}\left(  \mathbf{\xi}_{j}^{T}{{\mathbf{H}%
}_{2,j}}\right)  \boldsymbol{\Lambda}{_{j}}{{\left(  \mathbf{\xi}_{j}%
^{T}{{\mathbf{H}}_{2,j}}\right)  }^{T}.} \label{65}%
\end{equation}

According to (\ref{63}) and (\ref{65}), one has ${{d}_{j}}\leq d_{j}^{c}$.
Then $\rho\left(  \mathbf{d},\partial\Omega\right)  \leq\rho\left(
\mathbf{u}{_{c}},\partial\Omega\right)  $ according to (\ref{62}) and
(\ref{64}). Then, $\rho\left(  \mathbf{u}{_{c}},\partial\Omega\right)  $ is
the maximum ACAI of system (\ref{distrubance_model}). $\square$

In the following, $\rho\left(  \mathbf{u}{_{c}},\partial\Omega\right)  $ will
be used to define the DoC of the multicopter system:

\textbf{Definition 1 (Degree of Controllability for Multicopters)}. The degree
of controllability for the multicopter system in (\ref{distrubance_model}) is
defined as
\begin{equation}
\sigma\triangleq\frac{{\rho\left(  {{\mathbf{d}},\partial\Omega}\right)  }%
}{{\rho\left(  \mathbf{u}{_{c},\partial\Omega}\right)  }} \label{26}%
\end{equation}
where $\rho\left(  \mathbf{d},\partial\Omega\right)  $ is the ACAI of the
multicopter system.

From \emph{Definition 1}, one can see that
\begin{equation}
\rho\left(  \mathbf{d},\partial\Omega\right)  =\sigma\centerdot\rho\left(
\mathbf{u}{_{c}},\partial\Omega\right)  . \label{28}%
\end{equation}
Here, $\sigma$ shows the impact of the disturbance $\mathbf{d}$ on the virtual
system, where the disturbance in the system (\ref{distrubance_model})
satisfies $\mathbf{d}=\mathbf{u}{_{c}}$. Considering the basic dynamics, if
$\mathbf{d}=\mathbf{0}$ when the multicopter is hovering, then there is no
control margin to land the multicopter as the propulsors can only provide
upwards thrust. Similarly, if $\mathbf{d}=2\mathbf{u}{_{c}}$, namely all the
propulsors are providing the maximum thrust, then there is no control margin
to lift the multicopter anymore.

According to (\ref{22}), $\rho\left(  \mathbf{d},\partial\Omega\right)  \leq0$
if the multicopter system is uncontrollable. For the sake of simplicity, let
\begin{equation}
\sigma=0,\ \text{if}\ \rho\left(  \mathbf{d},\partial\Omega\right)  \leq0.
\label{27}%
\end{equation}
Then $\sigma=0$ when the multicopter system in (\ref{distrubance_model}) is
uncontrollable. According to \emph{Lemma 1}, (\ref{26}) and (\ref{27}), the
following theorem holds:

\textbf{Theorem 2}. For the system in (\ref{distrubance_model}), the DoC
satisfies $\sigma\in\left[  0,1\right]  $.

\subsubsection{Definition of the Control Performance Index}

Although the DoC $\sigma$ shows how controllable the system is,
controllability does not imply stability. In practice, people concern
stability more than controllability of a multicopter flying in the air. As a
result, a good index should satisfy: i) it is nonpositive if the system is
unstable; ii) it is positive if the system is stable; iii) the larger the
index value is, the more stable is the system. To show the stability
performance of the flying vehicle, this paper will define a new control
performance index based on the DoC $\sigma$.

Suppose that $\mathbf{d}\in U_{d}\triangleq\left\{  \mathbf{d}{|}d_{i}%
\in{\left[  d_{i,\min}{,}d_{i,\max}\right]  ,i=}1,\cdots,{n}\right\}  $. Then
a new constraint set $U_{\sigma_{0}}$ is defined as%
\begin{equation}
U_{\sigma_{0}}\triangleq\left\{  \mathbf{d}|\sigma=\frac{{\rho\left(
{{\mathbf{d}},\partial\Omega}\right)  }}{{\rho\left(  \mathbf{u}{_{c}%
,\partial\Omega}\right)  }}\geq\sigma_{0},\mathbf{d}\in U_{d}\right\}  .
\label{def_ud}%
\end{equation}
where $U_{\sigma_{0}}$ contains all the disturbances satisfying ${\rho\left(
{{\mathbf{d}},\partial\Omega}\right)  }\geq\sigma_{0}{\rho\left(
\mathbf{u}{_{c},\partial\Omega}\right)  }$. Given a controller $\mathbf{u}%
=\mathbf{u}\left(  \mathbf{x},t\right)  $, a large enough disturbance will
make closed-loop of system (\ref{distrubance_model}) unstable. Then a
definition of control performance threshold (CPT) ${{\sigma}_{th}}$ is given
as follows:

\textbf{Definition 2 (Control Performance Threshold)}. The CPT of the system
in (\ref{distrubance_model}) is defined as
\begin{equation}
{{\sigma}_{th}}\triangleq\underset{{{\sigma}_{s}}\in\left[  0,1\right]  }%
{\inf}\left(  {{\sigma}_{s}}\right)  \label{29}%
\end{equation}
where the variable ${{\sigma}_{s}}$ satisfies the following condition: for a
given ${{\sigma}_{s}}$, the system (\ref{distrubance_model}) subject to
disturbance $\forall\mathbf{d}\in U_{\sigma_{0}=\sigma_{s}}$ (i.e., any
$\mathbf{d}$ satisfying ${\rho\left(  {{\mathbf{d}},\partial\Omega}\right)
}\geq{{\sigma}_{s}\rho\left(  \mathbf{u}{_{c},\partial\Omega}\right)  }$) is
stable under the controller $\mathbf{u}=\mathbf{u}\left(  \mathbf{x},t\right)
$.

Without loss of generality, the CPT ${{\sigma}_{th}<1}$ for the system
(\ref{distrubance_model}) controlled by a \emph{reasonable }and\emph{\ robust}
strategy $\mathbf{u}=\mathbf{u}\left(  \mathbf{x},t\right)  $. Suppose that
${{\sigma}_{th}}$ is determined for the control strategy $\mathbf{u}%
=\mathbf{u}\left(  \mathbf{x},t\right)  $. Then from the definition in
(\ref{29}) the closed-loop system will be stable if $\sigma\geq{{\sigma}_{th}%
}$ and unstable otherwise. In order to show the stability margin of the
closed-loop system intuitively, a control performance index is defined based
on \emph{Definition 1} and \emph{Definition 2}:

\textbf{Definition 3 (Control Performance Index)}. The CPI of the multicopter
system in (\ref{distrubance_model}) is defined as
\begin{equation}
S\triangleq\frac{\sigma-{{\sigma}_{th}}}{1-{{\sigma}_{th}}} \label{30}%
\end{equation}
where ${{\sigma}_{th}<1}$ is CPT of the closed-loop system in
(\ref{distrubance_model}) with the control strategy $\mathbf{u}=\mathbf{u}%
\left(  \mathbf{x},t\right)  $.

From \emph{Definition 3}, we say that the multicopter is safe if $S\geq0$ and
unsafe otherwise. As $\sigma\in\left[  0,1\right]  $, then one has
\begin{equation}
S\in\left[  \frac{-{{\sigma}_{th}}}{1-{{\sigma}_{th}}},1\right]  . \label{31}%
\end{equation}
Now the stability of the multicopter system can be indicated by the CPI $S$.

\subsection{Threshold Value Determination}

As mentioned above, there is a CPT ${{\sigma}_{th}}$ for the specified control
strategy $\mathbf{u}=\mathbf{u}\left(  \mathbf{x},t\right)  $, and the
closed-loop system with the control strategy is stable if $\sigma\geq{{\sigma
}_{th}}$. Although one may obtain the theoretical value of ${{\sigma}_{th}}$
if the explicit expression of the controller $\mathbf{u}=\mathbf{u}\left(
\mathbf{x},t\right)  $ is simple, it is hard to compute ${{\sigma}_{th}}$
theoretically because the controller is either complex or accessible only in
part in practice. In this paper, the CPT ${{\sigma}_{th}}$ is obtained through
numerical simulations and/or real flight experiments. By taking the lumped
disturbance $\mathbf{d}$ into account, the computing procedure is given by
\emph{Algorithm 1}.

\smallskip

\noindent\rule[0.25\baselineskip]{0.5\textwidth}{0.75pt}

{\small \noindent\textbf{Algorithm 1} Threshold value determination procedure
}

\noindent\rule[0.25\baselineskip]{0.5\textwidth}{0.5pt}

{\small \noindent\textbf{Step 1}: Generate the disturbance grid set
$\Xi\subset{{\mathbb{R}}^{{n}}}$ of the disturbance $\mathbf{d}\in U_{d}$. As
$\mathbf{d}\in U_{d}$, the constraint of $d_{i}$ can be obtained as $d_{i}%
\in\left[  d_{i,\min},d_{i,\max}\right]  $ where $i=1,\cdots,{n}$ and
$d_{i,\min},d_{i,\max}$ are the minimum and maximum value of $d_{i}$
respectively. Suppose that $\left[  d_{i,\min},d_{i,\max}\right]  $ is divided
into $n_{d}$ grid points, then $U_{d}$ changes to $\Xi\subset{{\mathbb{R}%
}^{{n}}}$ with $n_{d}^{n}$ points. }

{\small \noindent\textbf{Step 2}: Compute the ACAI of the multicopter system
(\ref{distrubance_model}) corresponding to each disturbance grid points in
$\Xi$. }

{\small \noindent\textbf{Step 3}: Compute the DoC $\sigma$ of each disturbance
grid point in $\Xi$, and the results are denoted by set $\Lambda$. }

{\small \noindent\textbf{Step 4}: Let $k=0$, and $\Delta\sigma\in\left(
0,1\right]  $. }

{\small \noindent\textbf{Step 5}: Let $k=k+1$. If $k\Delta\sigma>1$, go to
\emph{Step 8}. }

{\small \noindent\textbf{Step 6}: Check the stability of the specified control
strategy $\mathbf{u}=\mathbf{u}\left(  \mathbf{x},t\right)  $ for all the
disturbance grid points satisfying $1-k\Delta\sigma\leq\sigma<1-\left(
k-1\right)  \Delta\sigma$, which is denoted by {$\Xi$}${_{k}}$. }

{\small \noindent\textbf{Step 7}: If the closed-loop system with the control
strategy $\mathbf{u}=\mathbf{u}\left(  \mathbf{x},t\right)  $ is stable under
all the specified disturbance grid points in {$\Xi$}${_{k}}$, go to \emph{Step
5}. If the control strategy is unstable under any of the specified disturbance
grid point in {$\Xi$}${_{k}}$, go to \emph{Step 8}. }

{\small \noindent\textbf{Step 8}: The CPT is obtained as ${{\sigma}_{th}%
}=1-\left(  k-1\right)  \Delta\sigma$. }

\noindent\rule[0.25\baselineskip]{0.5\textwidth}{0.5pt}

It should be pointed out that the nonlinear dynamics shown in equation
(\ref{nonlinear}) is applied in the simulations while the linear model
(\ref{distrubance_model}) is only used to compute the DoC of the multicopter.
From the above, the larger the value of $n_{d}$ is and the smaller the value
of $\Delta\sigma$ is, the more accurate is the threshold ${{\sigma}_{th}}$. In
practice, the CPT needs to be checked by real flight experiments. Fortunately,
there is no need to check all the disturbance grid points that satisfy
$\sigma\geq{{\sigma}_{th}}$. If the closed-loop system with the control
strategy $\mathbf{u}=\mathbf{u}\left(  \mathbf{x},t\right)  $ is stable with
all the disturbance grid points that satisfy ${{\sigma}_{th}}\leq\sigma
\leq{{\sigma}_{th}}+{{C}_{\sigma}}$, then the closed-loop system is always
stable if $\sigma>{{\sigma}_{th}}+{{C}_{\sigma}}$, where ${{C}_{\sigma}}$ is a
specified confidence value. Therefore, only the disturbance grid points that
satisfy ${{\sigma}_{th}}\leq\sigma\leq{{\sigma}_{th}}+{{C}_{\sigma}}$ need to
be tested by experiments.

\section{Application of the Control Performance Index}

In this section, the proposed CPI is used for a switching control framework
for multicopters, where the index is used to show how safe the vehicle is. The
on-ground pilots can decide to continue or abort the mission based on this
index. Besides, the index can guide the on-board autopilot to switch from
nominal control strategies to degraded control strategies for safe landing.

\subsection{Switching Control Framework}

\begin{figure}[t]
\begin{center}
\includegraphics[
scale=0.6 ]{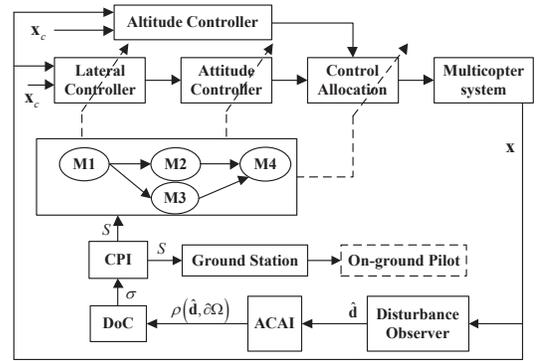} \vspace{-1.5em}
\end{center}
\caption{Switching control framework}%
\label{Fig_4}%
\end{figure}

The diagram of the switching control framework is shown in Fig.\ref{Fig_4}.
Practically, the reference signal $\mathbf{x}_{c}$ is given by the top level
guidance module or the pilots on the ground. As shown in Fig.\ref{Fig_4}, a
typical multicopter usually has lateral position controllers, altitude
controller, attitude controllers and a control allocation module. The vehicle
has the following three control modes: i) \emph{Mode M1,} where the lateral
dynamics and the basic dynamics are controlled by nominal lateral controller,
altitude controller and attitude controller; ii) \emph{Mode M2}, where the
lateral dynamics are given up and the basic dynamics are controlled by the
nominal altitude and attitude controllers; iii) \emph{Mode M3}, where the
lateral dynamics are controlled by nominal lateral controller, and the
altitude is controlled by nominal controller while the attitude is controlled
by degraded controllers; iv) \emph{Mode M4}, where the lateral dynamics is
given up, and the altitude is controlled by nominal controller while the
attitude is controlled by degraded controllers. 

There are many research on the nominal control of multicopters, see
\cite{mahony2012multirotor,tayebi2006attitude,lee2009feedback,wang2015nonlinear}
and the references therein. To make this paper more extensible, the nominal
control strategy in the framework is not specified. For the case that the
multicopter under severe off-nominal conditions is uncontrollable, a degraded
control strategy will be adopted. The papers
\cite{Freddi2011ifac,Akhtar2013,Lanzon2014jgcd,du2015jint,Mueller2015Andrea}
studied a relaxed hover solution for multicopters where the vehicle may rotate
at a constant velocity in hover, by giving up the control of the yaw angle
(the yaw states are ignored). Then, these strategies are now integrated with
an online estimator for the CPI, resulting in a switching control system that
is robust against off-nominal conditions. Besides, under this framework the
real-time control performance state of the multicopters can be sent to the
ground station so that the on-ground pilots can decide to continue or abort
the mission.

In practice, many kinds of disturbance observers (such as Kalman filter) can
be used to estimate the disturbance based on the dynamic model shown in
(\ref{distrubance_model}). If a Kalman filter is used to estimate the
disturbance, the estimated disturbance covariance can be used to obtain
confidence on $\mathbf{\hat{d}}$. The ACAI can be obtained according to the
computation procedure given in \cite{du2015jgcd} and the toolbox in
\cite{du2016toolbox}. Based on the ACAI, the CPI of the system are obtained
according to the results in Section III. In the following, a switching control
strategy based on the CPI is proposed.

\subsection{Switching Control Strategy Based on the Control Performance Index}

The CPI results not only tell the autopilots whether it is necessary to switch
to the degraded controllers or not but also tell the on-ground pilots the safe
state of the vehicle. If necessary, the on-ground pilots can land the vehicle
before LOC accidents occur. The CPI can be used in the following scenarios: i)
Before the mission starts, the on-ground pilot can evaluate the control
performance by a short time flight. Excessive payload and propulsor faults
will be checked out based on the CPI results. ii) In the case of high wind,
the on-ground pilots or the onboard autopilots will land the vehicle
immediately if the CPI approaches a value small enough before the vehicle
becomes unstable. iii) In the case of sudden severe conditions, the on-ground
pilot could not predict the safe state based on the CPI history and can not
make a safety decision. At this time, the vehicle will try to land
automatically in a degraded way or the vehicle will be LOC.

Denote the CPI of the lateral dynamics, the basic dynamics, and the degraded
system by $S_{l}$, $S_{b}$, and {$S_{d}$}, respectively. Then, the following
observations are obtained according to \emph{Definition 1}, \emph{Definition
2}, and \emph{Definition 3}: i) \textbf{Observation 1}. If $S_{l}>0$, and
$S_{b}>0$, then the multicopter is safe to continue the flight. ii)
\textbf{Observation 2}. If $S_{l}\leq0$, and $S_{b}>0$, then the lateral
dynamics is given up and only the basic dynamics of the vehicle is controlled
by the nominal control. iii) \textbf{Observation 3}. If $S_{l}>0$, and
$S_{b}\leq0$, then the lateral dynamics is controlled by the nominal control,
and the yaw states are given up. Then, the degraded control strategy is used
to land the vehicle safely. iv) \textbf{Observation 4}. If $S_{l}\leq0$,
$S_{b}\leq0$ and {$S_{d}$}${>0}$, then the lateral dynamics and the yaw states
are given up and only the degraded control strategy is used to land the
vehicle safely. As the details of the degraded control strategy are beyond the
scope of this paper, the readers are referred to
\cite{du2015jint,Mueller2015Andrea} for more information.

\subsection{Closed-loop Stability Statement}

According to Fig.1, the vehicle can only switch from M1 to M2/M3, and then
from M2/M3 to M4 for safety considerations. In practice, if the lumped
disturbance $\mathbf{d}$ makes the vehicle switch from M1 to M2/M3, then it
means that the flight conditions are not safe for the mission. If the
disturbance makes the vehicle switch from M2/M3 to M4, then it means that the
vehicle is in an ill-condition, and the vehicle should land immediately. From
the above, the chattering problem is prevented because the vehicle can only
switch from M1 to M2/M3, and then from M2/M3 to M4 unidirectionally. In the
future, the switch from M4 to M2/M3 can also be considered. In this case, the
average dwell time for each mode should be defined to ensure the stability of
the switched system \cite{hespanha1999stability,lin2009stability}.

\section{Simulation and Experiments}

To show the effectiveness of the proposed CPI, both numerical and experimental
results are given in this section. Concretely, a \emph{hexacopter} subject to
propulsor faults is used to show the effectiveness of the proposed CPI and
switching control framework. On the other hand, a number of real flight
experiments are carried out to show the effectiveness of the CPI based on a
\emph{quadcopter} platform. Under the proposed framework, the off-nominal
behaviors, such as additional payloads, propulsor degradation and unacceptable
wind disturbances, are all lumped as a disturbance. Therefore, for simplicity
and without loss of generality, only the propulsor degradation (in the
simulations) and additional payloads (in the experiments) are considered.

\subsection{Simulations and Results}

Here, a traditional hexacopter with symmetric configuration (see
\cite{du2015jgcd} for the detailed parameters of the hexacopter) is considered
to show the effectiveness of the proposed CPI and the switching control
framework. The simulation model of the hexacopter is constructed which
consists of three main modules: i) two control strategies: the nominal control
strategy and the degraded control strategy, ii) a real-time estimator to
obtain the CPI $S_{l}$, $S_{b}$ and {$S_{d}$}, iii) a switching control
strategy based on the CPI. In the simulation, the hexacopter is controlled to
1 meter above the ground (${{h}_{c}}=1$), and maintains the level state
(${{\phi}_{c}}={{\theta}_{c}}={{\psi}_{c}}=0$).

To compute the ACAI $\rho\left(  \mathbf{d},\partial\Omega\right)  $, a Kalman
filter is used to estimate the lumped disturbance $\mathbf{d}$. Based on the
estimated disturbance {{$\mathbf{\hat{d}}$}}, the value of the ACAI can be
computed according to the procedure presented in \cite{du2015jgcd}. Then, the
DoC $\sigma$ and the CPI $S_{b}$ can be computed based on the ACAI. Similarly,
the lumped disturbance of the degraded system can be estimated and the DoC
(denoted by ${\bar{\sigma}}$) and the CPI {$S_{d}$} can be computed. And the
CPI $S_{l}$ can also be computed in the similar way to $S_{b}$.

\subsubsection{Threshold Value Determination}

\begin{table}[ptb]
\caption{Threshold value determination for $\sigma$ ($N_{\text{total}}$ is the
total points number, $N_{\text{stable}}$ stable points number)}%
\label{normthreshold}
\centering%
\begin{tabular}
[c]{lccc}\hline\hline
$\sigma$ & $N_{\text{total}}$ & $N_{\text{stable}}$ & percentage\\\hline
1 & 3 & 3 & 100\%\\
\lbrack0.9,1) & 9 & 9 & 100\%\\
\lbrack0.8,0,9) & 90 & 90 & 100\%\\
\lbrack0.7,0,8) & 242 & 242 & 100\%\\
\lbrack0,6,0.7) & 478 & 478 & 100\%\\
\lbrack0.5,0.6) & 843 & 843 & 100\%\\
\lbrack0.4,0.5) & 1329 & 1329 & 100\%\\
\lbrack0.3,0.4) & 1865 & 1848 & 99\%\\
\lbrack0.2,0.3) & 2705 & 2380 & 88\%\\
\lbrack0.1,0.2) & 3190 & 2245 & 70\%\\
\lbrack0,0.1) & 183727 & 1544 & 0.1\%\\\hline\hline
\end{tabular}
\end{table}

According to the computing procedure for the threshold value in Section III.C,
we set $n_{d}=21$, $\Delta\sigma=0.1$, and simulations are performed and the
results are shown in Table \ref{normthreshold}. From Table \ref{normthreshold}%
, it can be seen that the system (\ref{distrubance_model}) controlled by the
nominal control strategy is always stable if $\sigma\geq0.4$. Similarly, the
degraded system controlled by a degraded control strategy is simulated and is
always stable if ${\bar{\sigma}}\geq0.4$ (details are omitted here). And the
lateral system controlled by the nominal control strategy is simulated and is
always stable if the DoC ${\sigma}_{l}\geq0.5$ (details are omitted here).
Then we get the threshold value of the considered hexacopter as follows%
\begin{equation}
{{\sigma}_{th}}=0.4,{{{\bar{\sigma}}}_{th}}=0.4,{{\sigma}_{l,th}}=0.5.
\label{55}%
\end{equation}

\subsubsection{Simulation Results}

\begin{figure}[t]
\begin{center}
\includegraphics[
scale=0.6]{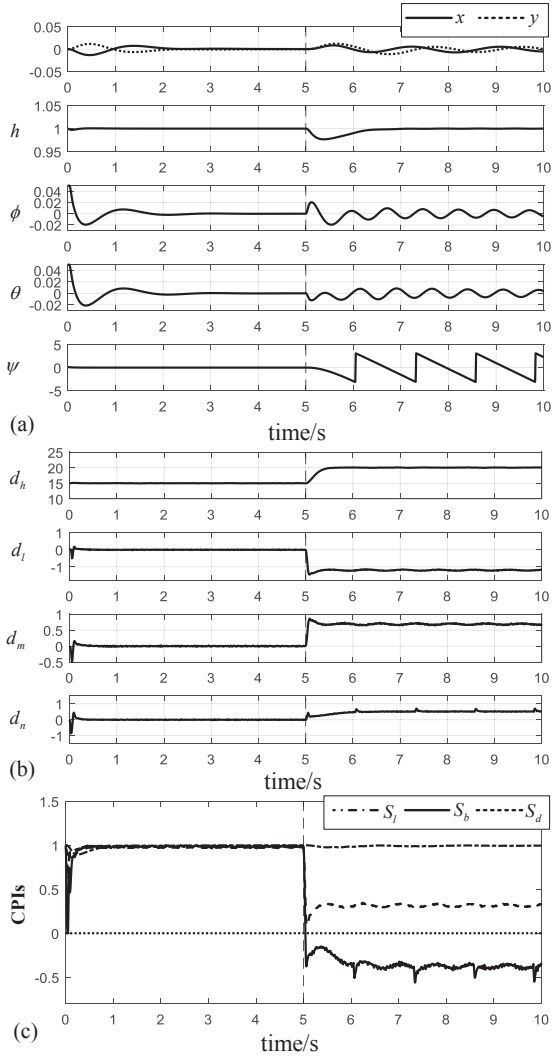} \vspace{-1.5em}
\end{center}
\caption{Simulation results: (a) Altitude and attitude states. (b) The
estimation of the lumped disturbance $\mathbf{\hat{d}}$. (c) The real-time
CPIs $S_{b},{S}_{d}$.}%
\label{Fig_sim}%
\end{figure}

In the simulation, the hexacopter is hovering with the altitude and the
roll-pitch-yaw angles under control. At time $t=5$s, propulsor 2 fails and the
degraded control strategy is switched to based on the switching control
methodology. The simulation results are shown in Fig.\ref{Fig_sim}. In
Fig.\ref{Fig_sim}(a), the real-time altitude and attitude information are
shown, where the multicopter is in \emph{Mode M1} when no faults occurred and
then switched to \emph{Mode M3} after propulsor 2 fails. The real-time
estimation of the lumped disturbance $\mathbf{\hat{d}}$ is shown in
Fig.\ref{Fig_sim}(b). The real-time CPIs $S_{l}$, $S_{b},${$S_{d}$} are shown
in Fig.\ref{Fig_sim}(c) from which it is observed that $S_{l}>0$, $S_{b}<{0}$
and {$S_{d}$}$>{0}$ after the failure of propulsor 2.

\begin{figure}[h]
\begin{center}
\includegraphics[
scale=0.6 ]{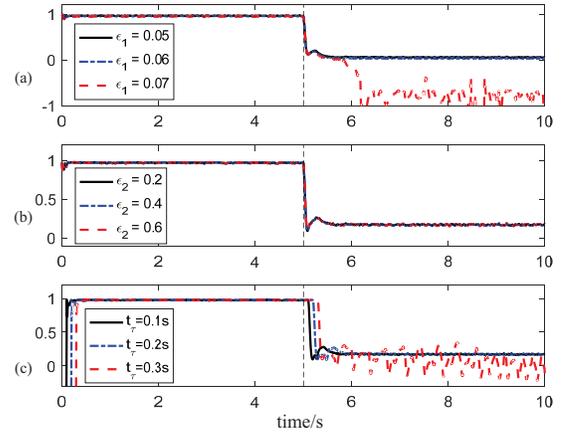} \vspace{-1.5em}
\end{center}
\caption{Effects of estimation bias, noise, and delays to the recovery
actions: (a) $S_{d}$ with bias introduced and $\epsilon_{2}=0$, $t_{\tau}=0$s.
(b) $S_{d}$ with noise introduced and $\epsilon_{1}=0$, $t_{\tau}=0$s. (c)
$S_{d}$ with delay introduced and $\epsilon_{1}=0$, $\epsilon_{2}=0$. }%
\label{robust}%
\end{figure}

To show how the uncertainties in the estimation process affect the
effectiveness of the recovery actions, estimation bias, different levels of
noise and estimation phase delays are introduced to the basic dynamics. Here,
the bias is denoted by $\mathbf{d}_{bias}$ and the time delay is denoted by
$t_{\tau}$. The standard deviation of the position and attitude sensor noises
are denoted by $\chi_{p}$ and $\chi_{a}$, respectively. The simulation results
are shown in Fig.\ref{robust}. In Fig.\ref{robust}(a), the bias $\mathbf{d}%
_{bias}=\epsilon_{1}\mathbf{d}_{0}$ where $\mathbf{d}_{0}\approx\lbrack21$
$-1.4$ $0.9$ $0.6]^{T}$ and $\epsilon_{1}$ is set to be 0.05, 0.06, and 0.07.
In Fig.\ref{robust}(b), the Kalman filter is designed based on given
measurement noise $\chi_{p}=0.1$ and $\chi_{a}=0.01$ while the simulated
measurement noises (denoted by $\chi_{p}^{\prime}$ and $\chi_{a}^{\prime}$)
are $\chi_{p}^{\prime}=\left(  1+\epsilon_{2}\right)  \chi_{p}$, $\chi
_{a}^{\prime}=\left(  1+\epsilon_{2}\right)  \chi_{a}$. Here, $\epsilon_{2}$
is set to be 0.2, 0.4, and 0.6. In Fig.\ref{robust}(c), different delays
($t_{\tau}$ is set to be 0.1s, 0.2s, 0.3s) are introduced to the disturbance
estimation. From the results shown in Fig.\ref{robust}: i) different levels of
noise do not affect the effectiveness of the recovery action, ii) estimation
bias will shift the estimated CPI and may make the recovery action fail, iii)
the delay term $t_{\tau}$ will delay the recovery action. However, if the
estimation bias and the delays are small enough, the recovery action is
effective.

According to the simulations results, it is shown that the switching control
framework based on the proposed CPI is effective. In the following,
experiments are carried out to show the effectiveness of the real-time CPI
estimator in the switching control framework.

\subsection{Experimental Results}

A quadcopter platform named Qball-X4 \cite{YMZ2013FDD} (a quadcopter developed
by Quanser) is used in the experiments to show that the CPI can be used to
monitor the performance of the vehicle. A group of PID controllers are offered
by the manufacturer of the Qball-X4 for altitude and attitude control purpose.
To get the CPT of Qball-X4, the Qball-X4 simulation model offered by the
Quanser Company is modified slightly and the threshold value determination
procedures are used. Here, the details are omitted and the CPT of the Qball-X4
is $\sigma_{th}=0.3993$. \begin{figure}[t]
\begin{center}
\includegraphics[
scale=0.6]{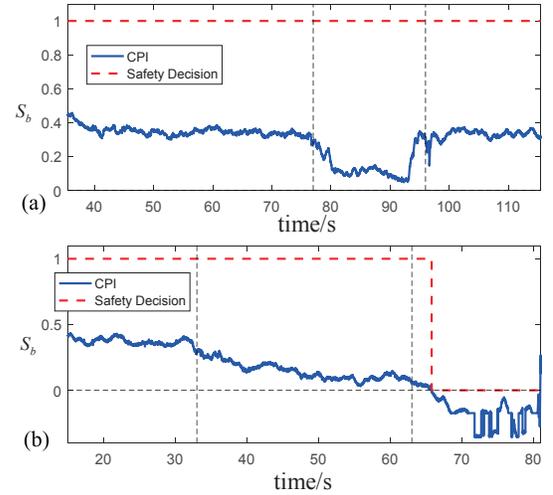} \vspace{-1.5em}
\end{center}
\caption{Experimental results: (a) A 100g weight was attached to the Qball-X4
at time $t=77$s and removed at time $t=96$s. (b) A 100g weight was attached to
the Qball-X4 at time $t=33$s and then a 50g weight was added at time $t=63$s.}%
\label{CPI}%
\end{figure}

In the experiments, different weights are attached to the same specified
place, and the real-time CPI will show the control performance of the
quadcopter. To verify the experimental results, the maximum weight allowed by
the quadcopter, which is $m_{\max}=126$g, is obtained by simulations. The main
purpose of the experiments here is to verify the effectiveness of the CPI
which is used to show the real-time control performance of the quadcopter and
guide the autopilot to make safety decisions (1 for safe and 0 for unsafe).
These experiments are recorded in the online video \cite{youtube} or
\cite{youku}, and the experimental results are shown in Fig.\ref{CPI}.

\subsubsection{Case 1: A 100g weight was attached to the Qball-X4}

Fig.\ref{CPI}(a) shows the experimental results where 100g weight is attached
to the specified place of the vehicle. The 100g weight was attached at time
$t=77$s, and the aircraft is still safe with the PID controllers. Then the
100g weight was removed at time $t=96$s. The CPI results in Fig.\ref{CPI}(a)
show that the Qball-X4 is always safe during the flight.

\subsubsection{Case 2: Totally 150g weight was attached to the Qball-X4}

However, in the second flight, totally 150g weight was attached to the same
place and the results are shown in Fig.\ref{CPI}(b). Firstly, 100g weight was
attached at time $t=33$s, the aircraft is safe. Then 50g weight was attached
to the same place as the 100g weight at time $t=63$s, and the CPI results in
Fig.\ref{CPI}(b) show that the Qball-X4 is unsafe after the 50g weight is
attached. From the video recording of this experiment, it is seen that
Qball-X4 is oscillating. The safety decision results are reasonable as the
\emph{maximum weight} allowed is 126g while totally 150g weight was attached.

\begin{figure}[tbh]
\begin{center}
\includegraphics[
scale=0.7]{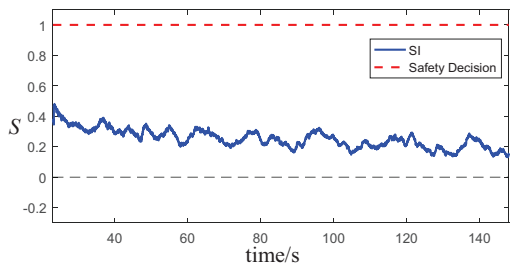} \vspace{-1.5em}
\end{center}
\caption{The effect of battery to $S$.}%
\label{battery}%
\end{figure}

From Fig.\ref{CPI}, one can see that the $S_{b}$ measurement seems to drift a
lot during a given experimental process. This is caused by decreased battery
voltage which is equivalent to an extra weight added to the vehicle.
Fig.\ref{battery} shows the results where no weight is added to the Qball-X4,
and one can see that the index $S_{b}$ is decreasing with the flight time.

From the above experiments, it can be seen that the CPI proposed by this paper
is practically effective. The CPI can be used to monitor the real-time control
performance of the multicopters and tell the users whether the vehicle is safe
to operate.

\section{Conclusions}

This paper studied the performance assessment problem of multicopters subject
to off-nominal conditions. Firstly, a new definition of Degree of
Controllability (DoC) was proposed for multicopters subject to control
constrains and off-nominal conditions to show the available control authority
of the vehicle. Then, a control performance index (CPI) was defined based on
the new DoC to reflect the control performance of the multicopters. A
step-by-step procedure was also provided to obtain the control performance
threshold (CPT) which would be used to compute the CPI. Besides, the proposed
CPI is used to guide the switching control of multicopters in a new switching
control framework. Finally, simulation and experimental results showed the
effectiveness of the switching control framework and the CPI proposed in this paper.



\appendices

\section*{Acknowledgments}

This work is supported by the National Natural Science Foundation of China
(Grant No. 61603014 and No. 61473012) and the China Postdoctoral Science
Foundation (Grant No. 2016M600895).

\ifCLASSOPTIONcaptionsoff
\newpage\fi

\end{document}